\documentclass[10pt]{article}

\usepackage{amsmath}
\usepackage{amssymb}
\usepackage{algorithm, algorithmic}

\usepackage{graphicx}

\usepackage{cite}

\usepackage{color}
\usepackage{subfigure}
\usepackage{bm}

\usepackage{float,amsfonts,amsxtra,amsthm}

\usepackage[labelfont=bf,labelsep=period,justification=raggedright]{caption}

\makeatletter
\renewcommand{\@biblabel}[1]{\quad#1.}
\makeatother

\date{}

\newcommand{\argmin}{\operatornamewithlimits{argmin}}

\newcommand{\sgn}{\operatornamewithlimits{sgn}}

\begin{document}

\begin{flushleft}
{\Large
\textbf{Inferring Regulatory Networks by Combining Perturbation Screens and Steady State Gene Expression Profiles}
}
\\
Ali Shojaie$^{1,\#}$,
Alexandra Jauhiainen$^{2,\#}$,
Michael Kallitsis$^{3,\#}$
George Michailidis$^{3,\ast}$
\\
\bf{1} Department of Biostatistics, University of Washington, Seattle, WA, USA\\
\bf{2} Department of Medical Epidemiology and Biostatistics, Karolinska Institutet, Stockholm, Sweden\\
\bf{3} Department of Statistics, University of Michigan, Ann Arbor, MI, USA\\
$\#$ These authors contributed equally to this work.\\
$\ast$ E-mail: gmichail@umich.edu
\end{flushleft}

\section*{Abstract}
Reconstructing transcriptional regulatory networks is an important task in functional genomics. Data obtained from 
experiments that perturb genes by knockouts or
RNA interference contain useful information for addressing this reconstruction problem. 
However, such data can be limited in size and/or are expensive to acquire.
On the other hand, observational data of the organism in steady state (e.g. wild-type) are more readily available, 
but their informational content is inadequate for the task at hand.
We develop a computational approach to appropriately utilize both data sources for estimating a regulatory network. 
The proposed approach is based on a three-step algorithm
to estimate the underlying directed but cyclic network, that uses as input both
perturbation screens and steady state gene expression data. In the first step, the algorithm determines 
causal orderings of the genes that are consistent with
the perturbation data, by combining an exhaustive search method with a fast heuristic that in turn 
couples a Monte Carlo technique with a fast search algorithm. In the
second step, for each obtained causal ordering, a regulatory network is estimated using a penalized likelihood based method, 
while in the third step a consensus network is
constructed from the highest scored ones. Extensive computational experiments
show that the algorithm performs well in reconstructing the underlying network and clearly 
outperforms competing approaches that rely only on a single data source.
Further, it is established that the algorithm produces a consistent estimate of the regulatory network.


\section*{Introduction}

The ability to reconstruct cellular networks plays an important role in our understanding of how genes interact with each other and the way information flows through them to regulate their expression levels. Such reconstructions heavily depend on the input data employed. The availability of data on the response of the cell to {\em perturbations} -either by knocking out or silencing genes- offers the possibility for improved network reconstructions and constitutes a key input in functional genomics. As pointed out in \cite{markowetz2007nested}, high-dimensional phenotypic profiles obtained from perturbation experiments in the form of expression data offer the potential for obtaining a comprehensive view of cellular functions, even though they exhibit a number of limitations as outlined in \cite{tresch2008structure}.
A key problem is the fact that perturbation experiments only provide indirect information on gene interactions, as explained below
\cite{klamt2010transwesd}. Further, inferring large scale cellular networks from perturbation data is computationally challenging, and only a limited number of computational tools have been developed to address it.
Some approaches are built on clustering of phenotypic profiles \cite{piano2002gene,ohya2005high}, which are based on the reasoning that functionally related genes should exhibit similar behavior under perturbations and hence cluster together.
A tailor-made approach for the problem of estimating networks from perturbation data is the nested effects models (NEMs)
\cite{markowetz2005bionfo,markowetz2007nested,tresch2008structure}. NEMs are a special class of graphical models originally introduced to uncover the hierarchies among transcription factors based on observations of affected genes. More recently, NEMs have been extended to reconstruct regulatory networks by taking advantage of the nested structure of the observed perturbation effects, where for computational efficiency purposes, triplets of genes are used to assemble the global regulatory network.
Extensions of this method that capture temporal effects by using perturbation time series measurements are described in \cite{anchang2009modeling, frohlich2010fast}. In response to the reconstruction problems presented in the DREAM challenges (see more information in the Results section), methods like feed-forward loop down ranking (FFLDR)
\cite{pinna2010} and a $t$-test based method coupled with ordinary differential equations to model temporal changes in expression data (Inferelator)\cite{greenfield2010infer} were also developed.

The computational difficulty of reconstructing a network from data, alluded to above, stems from the fact that in order to capture the regulatory interactions one should consider all possible orderings of genes in the network (so that parent nodes influence child ones) and score the resulting network structures accordingly. The computational complexity of identifying all
possible orderings of a set of nodes in a directed graph is exponential in the size of the graph. 
Hence, the approach based on nested effects models employs several heuristics for searching the space of orderings.
Similarly, \cite{friedman2003being} employs a Markov Chain Monte Carlo based search method and subsequent scoring of the resulting network structures.

Another set of approaches solely utilizes observational gene expression data that capture  
the system in steady state. A major technical tool for such reconstructions is graphical
models \cite{pearl2000caus} that encode a probability model over the genes through the underlying network. 
Over the last few years, a number of algorithms have been
proposed in the literature for the estimation (reconstruction) 
of primarily Gaussian graphical models under the assumption of a sparse underlying structure
(see \cite{shojaie2010penalized,Michailidis2013} for a discussion and references therein). The main shortcoming of these approaches is
that graphical models for observational data are mostly capable of identifying dependencies between genes, 
rather than causal relations representing
regulatory mechanisms. Further, the presence of more genes than available samples usually leads to very sparse reconstructions. 
It should be noted that
recent work is geared towards identifying causal effects from observational data by employing the concept of intervention calculus \cite{maathuis2010predicting}.
Also utilizing only steady state gene expression data is a method implemented in the PC-algorithm \cite{spirtes2000cpa} that 
starts from a complete
undirected graph and recursively sparsifies it based on conditional independence decisions; 
directionality can only be inferred for a subset of edges
(due to the issue of observational equivalence \cite{pearl2000caus}) and is added as a post-processing step.

Time course gene expression data have also been used to estimate gene regulatory networks, using two classes of models. In the first approach, called Granger causality, the predictive effect of genes on each other are used to estimate regulatory relationships among genes \cite{yamaguchi2007fmb, opgen2007lcn}. In the second approach, known as dynamic Bayesian networks (DBNs), the framework of Bayesian networks is extended to incorporate biological networks with feedback loops \cite{ong2002mrp, perrin2003gni,Michailidis2013}. More recently, penalization methods have been applied to improve the estimation of high-dimensional regulatory networks from small sample sizes \cite{fujita2007mge, mukhopadhyay2007cps, shojaie2010discovering,shojaie2011thresholding}.

\begin{figure}[!t] 
 \begin{center}
 \includegraphics[width=10.75cm]{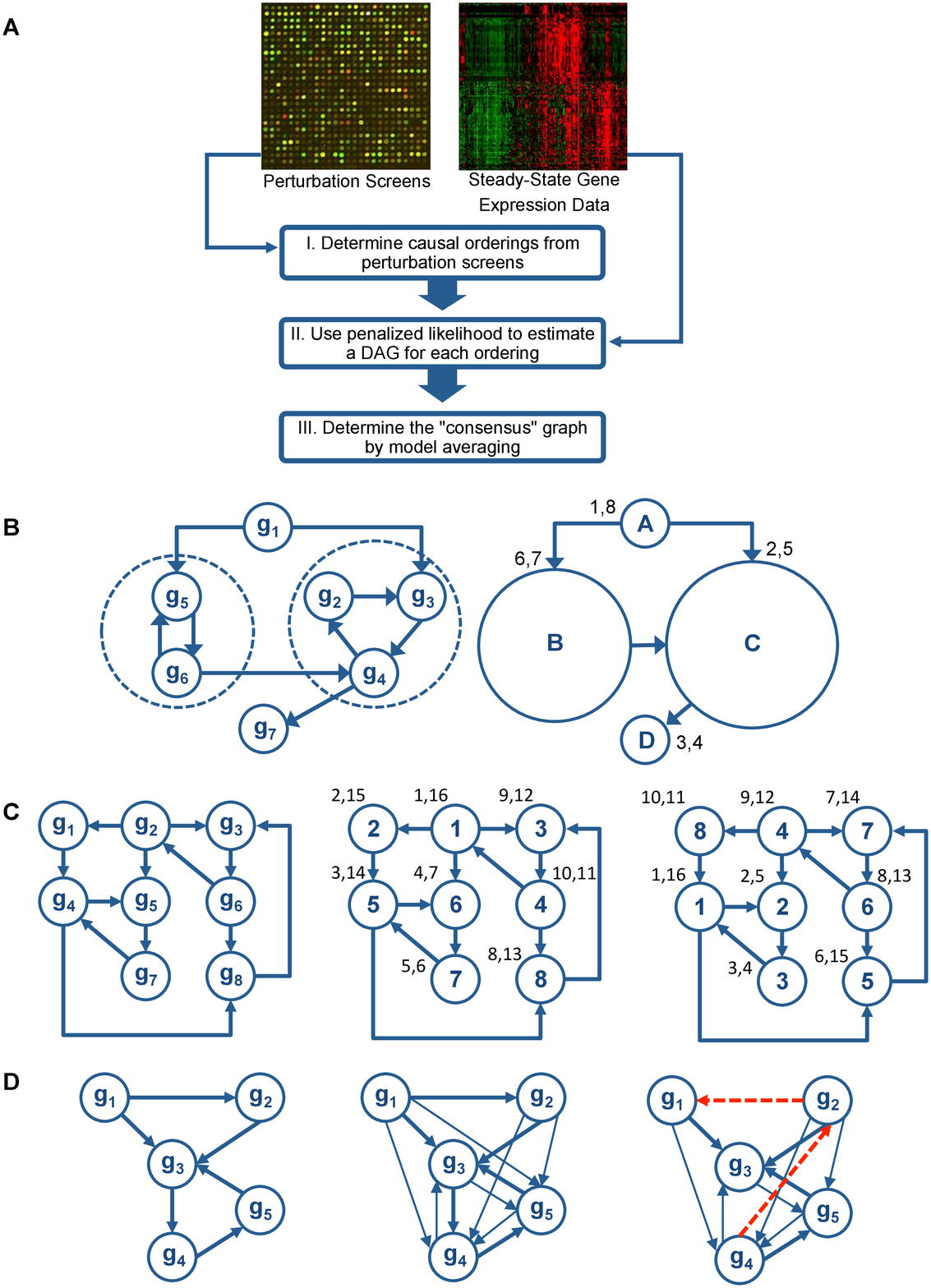}
 \end{center}
\caption{
{\bf Overview and details of the RIPE method.}
(A) Overview of the RIPE algorithm.
(B) First step to obtain a large set of causal orderings from a network with cycles. The network graph is decomposed into strongly connected components, so called super-nodes, (left) followed by a DFS on the strongly connected components (right). For example, the post-visit time of super-node A is 8, and thus A precedes all other nodes. The topological ordering of super-nodes is A $\prec$ B $\prec$ C $\prec$ D.
(C) Illustration of MC-DFS algorithm. Gene perturbation graph (left), DFS visit times for labeling \#1 (middle), and DFS visit times for labeling \#2 (right).
(D) Depiction of a small network to illustrate the influence graph and the predictors used in the penalized likelihood estimation procedure. 
The true regulatory network (left), the influence graph under no noise (middle), observed influence graph, with false positive and false negative edges (right). Edges in the regulatory network are shown in thick lines and additional edges in the influence graph are distinguished with narrow lines; red dash lines indicate false positives.
}
  \label{fig:ripe}
\end{figure}
The proposed computational approach in this study utilizes both gene expression data obtained from perturbation experiments,
and an independent expression data set reflecting steady state behavior of the cell. The main steps are summarized in Figure 1A.
Specifically, based on the perturbation phenotyping data, we obtain a (large) set of causal orderings of the genes through
a fast search algorithm which samples from the space of all possible orderings (see section Obtaining Causal Orderings from Perturbation Data
for the definition of a causal ordering and algorithmic details). These orderings correspond to the inherent layering of nodes of the graph.
Each set of orderings is then employed to obtain a directed acyclic regulatory graph/network (DAG) using the independent gene expression data set
through an extension of a fast penalized likelihood method introduced in \cite{shojaie2010penalized}. Further, the likelihood of every estimated graph is calculated. Finally, a {\em consensus} regulatory network (which can
very well contain cycles when the true network is cyclic) is obtained by averaging a small set of the most likely DAGs obtained.
The advantage of the proposed algorithm is that it utilizes both perturbation and steady state expression data, 
and once the set of gene orderings is
determined, a theoretically rigorous and computationally fast likelihood-based method for estimating the underlying network structure is used.
Further, causal orderings are determined by searching through the entire set of orderings consistent with the perturbation data, thus taking a global perspective,
as opposed to competing methods (e.g. the nested effects model or the PC-algorithm) that only utilize information pertaining to direct neighbors of nodes and then assembling the network from such local estimates.

\section*{Methods}
\label{sec:methods}
The proposed methodology has a number of additional advantages over competing methods.
First, the underlying assumption in the estimation of regulatory networks from perturbation data 
is that the structure of the network does not change in different knockout experiments, and therefore observations from these
experiments can be combined in order to estimate the underlying regulatory network. However, such an assumption is not fully valid:
biological systems and regulatory networks are highly robust \cite{stelling2004robustness, kitano2004biological}, which is believed
to be the result of redundant regulatory mechanisms, and knocking out a single gene may trigger an alternate regulatory pathway, resulting in a
different network structure. On the other hand, the main assumption of the methodology proposed in this paper is that the causal orderings of genes
remain stable in different perturbation experiments, which is significantly
less restrictive. Second, although the problem of reconstructing regulatory networks
is computationally NP-hard, the computational complexity of the approximate algorithm proposed in this paper is considerably lower than most methods of
network reconstructions directly from perturbation data. This is mainly due to the following facts:
(i) the space of possible orderings is smaller than the space of
graphs (even acyclic ones) \cite{friedman2003being}, and (ii) by employing the Monte Carlo sampling framework over the space of orderings, the approximate algorithm offers a
tractable alternative in high dimensional settings.
Our extensive experimental results indicate that the algorithm does not require an exhaustive search over the space of orderings, and a much smaller set of
orderings often results in significant improvements over competing methods.
Finally, it is known that perturbation data in the form of single gene knockouts do not provide sufficient information for 
estimation of regulatory networks and, in theory, all possible knockout combinations may be needed to fully discover a 
regulatory network. On the other hand, the indirect information in perturbation experiments is sufficient for estimating 
the causal orderings (see the subsection Obtaining Causal Orderings from Perturbation Data under Methods).
Therefore, by breaking down the network reconstruction problem into three steps, not only can we achieve 
better computational complexity, but we can also
transform a non-tractable problem into a sequence of tractable sub-problems. 
Further, the proposed approach offers the possibility to gain insight into
the informational contributions of the two data sources used, and offers 
improved performance through systematic integration of the two sources of data.

\subsection*{Method Overview}\label{model_1}
We introduce next a three-step algorithm, called \emph{RIPE} (standing for Regulatory network Inference from joint Perturbation and Expression data), which
incorporates both perturbation screens from knockout/knockdown experiments, as well as gene expression data usually reflecting steady
state behavior of the cell, in
order to address the problem of estimation of regulatory networks. We adopt the term {\em steady-state} data in our presentation for wild-type data.
The data obtained from perturbation (knock-out/down) experiments are referred to as {\em perturbation} data. Note though that in
practice, the actual ``lab measurements'' of the knockout effects are also considered in an equilibrium state. However, for presentation clarity, we differentiate the two above-mentioned sources of data. The main steps for RIPE are illustrated in the flowchart of Figure 1A. 
In the first step, the data from the perturbation screens are employed to obtain a large collection of causal orderings. 
In the second step, each causal
ordering is used in conjunction with the steady state data to obtain an estimated regulatory network through a penalized likelihood approach.
Finally, in the third step a consensus network is constructed from the best networks obtained in the second step. A detailed description of these three steps follows.

\subsection*{Obtaining Causal Orderings from Perturbation Data}\label{model_2}
\subsubsection*{Estimating the Influence Matrix from Perturbation Data}
Let $\mathcal{P}$ denote the {\em binary influence matrix} of size $k\times p$, 
with $p$ representing the number of genes under study and $k$ the number of
single genes that are knocked out/down\footnote{In general, the RIPE algorithm requires ``perturbation'' of individual genes. 
These perturbations can be in
the form of knockouts and/or knockdowns as illustrated in the the section DREAM4 Challenge under Results. 
Thus, all references to knockout experiments
throughout the text also include single gene knockdowns.}. 
Each knockout experiment is repeated $n_i$ times, and captures the effect of the knockout gene on
the remaining genes. Also required are observations on the unperturbed network (repeated $n_0$ times), 
to determine the baseline expression of the $p$ genes.

An entry in the influence matrix $\mathcal{P}(i,j)=1$ if the knockout experiment for gene $i$ is affecting gene $j$, and 0 otherwise. 
To assess whether an entry
is non-zero, meaning that gene $j$ is differentially expressed in the knockout experiment of gene $i$ compared to its baseline expression, we can use e.g. a (moderated) 2-sample $t$-test. Applying different cutoffs to the $p$-values obtained from such an analysis results in different number of non-zero entries in $\mathcal{P}$. In Section Determining the Influence Matrix under Results, we describe a systematic procedure for determining the appropriate $p$-value cutoff. It is therefore clear that as the quality of the perturbation experiments improves, 
and/or number of replicates for each perturbation experiments increases,
fewer false positive/negative edges are present in the $\mathcal{P}$. 
Well-conducted perturbation experiments with $n_i$ in the range of 2 to 5 replicates often provide the required level of accuracy.

From matrix $\mathcal{P}$ one can obtain the \emph{directed influence graph} $G_{\mathcal{P}}(V, E)$ of $\mathcal{P}$, 
where $V$ is the set of vertices
and $E$ the set of the graph edges. The graph $G_\mathcal{P}$ contains an edge from node $i$ to node $j$ if the corresponding entry $\mathcal{P}(i,j)=1$.

It is worth noting that the RIPE algorithm can also be used in the case where $k < p$, which is often 
of interest in the setting of regulatory networks.
In this setting $k$ is the number of transcription factors (TFs), and $p$ the total number of genes under study. Assuming that
there are no directed edges from target genes to TFs, causal orderings are then found only from TFs to other TFs as well as target genes.
To simplify the exposition of statistical models, we discuss the details of the algorithm in the setting where $k = p$, and defer 
the case of $k < p$ to Section Regulatory Network in Yeast, where we illustrate the application of RIPE in such a setting by estimating the gene regulatory network of yeast with $k = 269$ perturbations on transcription factors, and $p = 6051$ total genes.

\subsubsection*{Obtaining Causal Orderings based on the Influence Matrix}
A critical step in the proposed approach is to obtain  causal orderings of the genes from the influence matrix $\mathcal{P}$.
In case the influence matrix $\mathcal{P}$ describes ``acyclic" causal effects (i.e, the influence graph of $\mathcal{P}$ is acyclic), \emph{a single}
ordering of the underlying graph would suffice. However, most likely the perturbation matrix contains cyclic causal effects,  
due to the presence of feedback mechanisms in the regulatory network. Further, perturbation experiments usually yield noisy data, 
which could also result in
cycles in the influence matrix,
even if the underlying network is acyclic. Hence, one usually deals with an influence matrix whose underlying graph contains cycles
and an individual causal ordering is not sufficient. We discuss next how one can obtain a \emph{set} of causal orderings from the
influence graph $G_{\mathcal{P}}$. 

We start by providing key definitions for a \emph{linear ordering} of a set, \emph{topological ordering} of an acyclic directed graph, and \emph{causal ordering} of a directed graph.
A \emph{linear ordering} of the elements of a set is a collection of \emph{ordered pairs} of the set elements, 
such that the ordering of each pair satisfies a certain criterion.  In graph theory, an example of a linear ordering is a topological ordering.
A \emph{topological ordering} (known also as topological sort~\cite{Dasgupta:2006:ALG:1177299})
of a directed acyclic graph $G_{\mathcal{P}}$ comprising of $p$ nodes, corresponds to a linear list of the form
$(x_1, x_2, \ldots, x_p)$ of the nodes $\{1,\ldots, p\}$, with $x_l$ denoting the label of the node in the $l$-th position in the ordered list.
The ordering adheres to all partial relations $i \prec j$ implied by the graph $G_{\mathcal{P}}$,  where the relation $i \prec j$ is
interpreted as ``node $i$ precedes node $j$", i.e. there is an acyclic path from node $i$ to node $j$.
We define the \emph{causal ordering} of a directed graph to be the linear listing of nodes that corresponds to a valid
\emph{depth-first search} traversal of the graph, defined next. Before proceeding to the description of our algorithms, we
want to emphasize a distinction between the terms \emph{topological sort} and \emph{causal ordering}.
In acyclic graphs, a topological sorting is a special case of a causal ordering. In other words,
the former refers to graphs with no cycles; the latter refers to linear orderings of causal effects
induced by the influence graph, and are obtained from graphs that potentially have cycles.
The difference is that in the latter ordering, not all partial relations of the form $i \prec j$ hold.

A standard method in graph theory for obtaining a topological ordering of an acyclic graph is by employing
the depth-first search (DFS) algorithm \cite{citeulike:2204443,Dasgupta:2006:ALG:1177299} which is outlined in Algorithm~\ref{dfs}.
When the exact same algorithm is utilized in graphs with cycles, a causal ordering is obtained.
DFS is a graph traversal algorithm. It ``searches'' the graph by traversing it ``in-depth''. This means that when a node is discovered,
the algorithm will continue searching for undiscovered nodes adjacent to the current one.
When the algorithm reaches a ``dead-end'', it backtracks until it finds previously visited nodes
that have undiscovered neighboring nodes, and if this fails as well, it proceeds to new nodes not yet visited. The complete details of
the algorithm are shown in Algorithm~\ref{dfs}, where a recursive implementation of the method is given (similar to the one in~\cite{Dasgupta:2006:ALG:1177299}).
Note that the algorithm saves the time of the first discovery of the node (pre-visit time) and the time of final departure of the node (post-visit time).
Final departure for, say, node $i$ occurs when there are no paths starting from $i$ and leading to undiscovered nodes.
The ordering is readily acquired by a descending sort of the post-visit times, as shown in the right panel of Figure~1B. 

For cyclic graphs, a mere application of DFS does not produce sufficient information to help us reconstruct the regulatory network.
We would like to obtain as many causal orderings as possible, and evaluate each individual causal ordering using our penalized
likelihood method described in the next section.
To tackle the problem of obtaining a large set of causal
orderings in graphs with cycles, we need to introduce the following two steps.
\begin{itemize}
\item
In step 1, we \emph{decompose} the graph into its connected components. Specifically, a {\em strongly connected component}~\cite{Dasgupta:2006:ALG:1177299}
is a subgraph, such that there exists a path from every node to any other node in the subgraph. Hence, if we collapse each strongly connected component
into a single super-node, the resulting graph is a DAG (see Figure 1B, left panel). 
We then produce a \emph{topological sorting} of the super-nodes. Note that since the super-nodes form a DAG, a mere topological sort is sufficient (see Figure~1B, right panel).
\item
In step 2, we produce a set of \emph{causal orderings} for each super-node.
Recall that in the presence of cycles, a causal ordering is not a topological ordering; instead, it
is a linear listing of nodes that arises from a graph traversal using DFS. We propose two methods to obtain a set of causal orderings,
each described in the sequel. The first involves an exhaustive DFS search  and can be employed on strong components with relatively few nodes;
the other couples DFS with Monte Carlo sampling
and is suitable for large size strong components.
\end{itemize}
After completion of the above steps we combine the set of orderings corresponding to each super-node and
thus we obtain the ``universe" of causal orderings imposed by the influence matrix $\mathcal{P}$. For the network graph depicted in Figure~1B,
the universe of orderings includes $1 \prec \{5 \prec 6\} \prec \{2 \prec 3 \prec 4\} \prec 7$,
$1 \prec \{5 \prec 6\} \prec \{4 \prec 2 \prec 3\} \prec 7$, $1 \prec \{5 \prec 6\} \prec \{3 \prec 4 \prec 2\} \prec 7$,
$1 \prec \{6 \prec 5\} \prec \{2 \prec 3 \prec 4\} \prec 7$,
$1 \prec \{6 \prec 5\} \prec \{4 \prec 2 \prec 3\} \prec 7$, $1 \prec \{6 \prec 5\} \prec \{3 \prec 4 \prec 2\} \prec 7$.

In the exhaustive DFS approach, we modify and extend Knuth's backtracking algorithm \cite{DBLP:journals/ipl/KnuthS74} that was proposed for
generating all topological sorting arrangements of a DAG. The exhaustive search procedure is
initiated at every node of the strongly connected component (thus, a parallel/concurrent implementation of this is feasible).
This ensures that all possible causal orderings will be considered.
Now, suppose that our search method has just discovered a new node, say node $j$.
A key idea that ensures that all paths initiated from that node will be accounted, is to save all adjacent nodes of
the newly visited node in a circular list. Suppose the list of node $j$ contains nodes $\{i_1, i_2,..., i_k\}$.
Then, DFS  proceeds exactly as
discussed above, i.e., it traverses the graph ``in-depth'', starting from $i_1$.
After all paths originating from node $i_1$ are visited, the algorithms backtracks to $j$, and then the alternative paths will be explored by consulting our circular list (i.e., paths starting at $i_2$, etc).
However, exhaustively searching via backtracking has exponential complexity due to the huge number of
combinations of paths that DFS can take. This makes the method practically infeasible for 
relatively large strongly connected components (e.g. larger than $\sim 10$). Nevertheless, it represents a useful tool for obtaining the universe of orderings in small size components.

\begin{algorithm}[t]
\caption{Algorithm for DFS}
\label{dfs}
\begin{algorithmic}
\REQUIRE Graph $G(V, E)$
\ENSURE Arrays \emph{pre}, \emph{post} of pre-visit and post-visit times
\STATE clock= 1 \COMMENT{Graph traversal counter}
\FOR{all $v \in V$}
\STATE visited($v$) = false
\ENDFOR

\FOR{all $v \in V$}
\IF {not visited($v$)}
  \STATE explore($v$)
\ENDIF
\ENDFOR
\STATE
\STATE Procedure \textbf{explore}($v$)  \COMMENT{Find all nodes reachable from $v$}
\STATE visited($v$) = true
\STATE pre [$v$] = clock \COMMENT{Mark the time of first discovery of $v$}
\STATE clock = clock + 1
\FOR{each edge $(v,u) \in E$}
\IF {not visited($u$)}
  \STATE explore($u$)
\ENDIF
\ENDFOR
\STATE post [$v$] = clock  \COMMENT{Mark the time of final departure from $v$}
\STATE clock = clock + 1
\end{algorithmic}
\end{algorithm}

For large size strong components, we develop a \emph{fast approximation algorithm}, named \emph{MC-DFS}, that incorporates ideas from Monte Carlo sampling techniques.
MC-DFS consists of two simple steps: first, it employs a random labeling of the graph nodes; then, it runs DFS based on the current labeling.
The workings of the MC-DFS heuristic algorithm can be best demonstrated -- for $m=2$ label permutations -- with the example of Figure~1C. 
Suppose that the influence graph given by matrix $\mathcal{P}$ is the one depicted on the leftmost panel of the figure.
The middle panel depicts labeling $1$ and the rightmost panel shows labeling $2$. The DFS post- and pre-visit times for both cases
are as shown in the figure. Given labeling $1$, DFS produces the following causal ordering: G2 $\prec$ G1 $\prec$ G4 $\prec$ G8 $\prec$ G3 $\prec$ G6 $\prec$ G5 $\prec$ G7.
Given labeling $2$, DFS produces: G4 $\prec$ G8 $\prec$  G3 $\prec$ G6 $\prec$ G2 $\prec$ G1 $\prec$ G5 $\prec$ G7.
Thus, a large set of random permutations of the node labels, followed by an application of the DFS algorithm allows us to
obtain a significant number of causal orderings and efficiently
sample the space of all possible orderings. Obviously, the quality of the sampled space of orderings depends on the number of label permutations used
(for some empirical assessment see the Results and Discussion sections). The complexity of DFS is $O(|V|+|E|)$ for a graph with $|V|$ nodes and $|E|$ edges.
Thus, the total complexity of MC-DFS for a strong component of  $r$ nodes and $e$ edges is $O(m r + m e)$, should one decide to generate $m$ permutations.

\subsection*{Estimation of Network Structure Using Gene Expression Data}\label{model_3}
As mentioned above, a gene regulatory network can be represented by a directed graph, whose adjacency matrix is denoted by $A$.
The element $A_{j,i}=1$ if gene $i$ is directly regulated by gene $j$, and $A_{j,i}=0$ otherwise.
In the setting of graphical models \cite{pearl2000caus}, the nodes of the graph represent random variables $X_1, \ldots, X_p$ and the edges capture associations between them.

It was shown in \cite{shojaie2010penalized}, that if the underlying network is a DAG and an ordering of its nodes is {\em known},
then estimating  the network reduces to estimating its skeleton, a significantly simpler computational problem.
Specifically, the causal effects of random variables in a DAG can be explained using \emph{structural equation models} (SEM) \cite{pearl2000caus}. In the setting where the data are normally distributed, SEM's can be represented based on linear functions explaining the relationship between each node and the set of its parents in the network:
\begin{equation}\label{eqnSEMlin}
    X_i = \sum_{j \in \mathrm{pa}_i}{\theta_{ji} X_j} + Z_{i}, \hspace{0.5cm} i=1, \ldots, p
\end{equation}
Here, $\mathrm{pa}_i$ denotes the set of \emph{parents} of node $i$, and $Z_i$'s are latent variables representing the variation in each node unexplained by its parents (for normally distributed data, $Z_i \sim \mathcal{N}(0, \sigma^2)$). Finally, the coefficients $\theta_{ji}$ for the linear regression model represent the \emph{effect} of gene $j$ on $i$ for $j \in \mathrm{pa}_i$.

In the case of cyclic graphs, the above SEM representation for DAGs is not directly applicable. However, for each causal ordering from the previous section,
the nodes of the graph can be reordered to obtain a DAG. Hence, using the above representation, the problem of estimating the structure of the DAGs corresponding
to a causal ordering of nodes, say $o$, can be posed as a penalized likelihood estimation problem as shown in \cite{shojaie2010penalized}.
In particular, let $\mathcal{X}$ be the $n \times p$ matrix of gene expression data, whose columns have been re-arranged according to the causal
ordering $o$, and denote by $\mathcal{X}_{I}$ the submatrix obtained by columns of $\mathcal{X}$ indexed by set $I$. Then as shown in \cite{shojaie2010penalized},
the estimate of the adjacency matrix of DAGs under the general weighted lasso (or $\ell_1$) penalty, is found by solving the following $\ell_1$-regularized
least squares problems for $i=2, \ldots, p$
\begin{equation} \label{DAGest_lasso}
       \hat{A}_{1:i-1,i} = \argmin_{ \bm{\theta} \in \mathbb{R}^{i-1} } { \Big\{
              n^{-1} \| \mathcal{X}_{i} - \bm{\theta}^T\mathcal{X}_{1:i-1} \|_2^2  +
              \lambda_i \sum_{j=1}^{i-1}{|\theta_j| w_{ji}} \Big\}  }
\end{equation}
\noindent where $A_{1:i-1,i}$ denotes the first $i-1$ elements of the $i$th column of the adjacency matrix and $\lambda_i$ is the tuning parameter for each lasso regression problem. In Section Choice of Parameters and Properties of the Algorithm, we discuss the choice of this tuning parameter. Finally, $w_{ji}$ represents the weights of the lasso method; for the regular lasso penalty used here, $w_{ji} \equiv 1$.

In the RIPE algorithm, a natural extension of \cite{shojaie2010penalized} is to use the above penalized likelihood estimation framework in order to estimate a DAG for each ordering $o \in \mathcal{O}$, where $\mathcal{O}$ is the set of orderings found from the perturbation data.
However, the perturbation data provide additional information regarding the influence of the genes in the network. In particular, in the absence of noise, the set of parents of each gene in the regulatory network are a subset of the set of
parents in the influence graph. Using this observation, we generalize the penalized estimation problem in \eqref{DAGest_lasso} to limit the set of variables in each penalized regression to those of the parents of node $i$ in the influence graph, consistent with each ordering, which equates the set of all \textit{ancestors} of $i$ in the regulatory network.
 In other words, for each ordering $o$, the set of edges pointing to each gene in the regulatory graph is estimated by solving the following $\ell_1$-regularized regression (lasso) problem:
\begin{equation} \label{DAGest_lasso_const}
       \argmin_{ \bm{\theta} \in \mathbb{R}^{ |J_o| } } { \Big\{
              n^{-1} \| \mathcal{X}_{i} -  \bm{\theta}^T \mathcal{X}_{J_o} \|_2^2 + \lambda_i \| \bm{\theta} \|_1 \Big\}  }
\end{equation}
where $J_o \equiv {\rm pa}_i^{\mathcal{P}_o}$ denotes the set of parents of $i$ in the influence graph consistent with ordering $o$ and $\|\bm{\theta}\|_1$ is the $\ell_1$ norm of $\bm{\theta}$.

To illustrate the optimization problem for estimation of DAGs in the second step of RIPE, consider the regulatory network in the left panel of Figure~1D. 
The middle panel of the figure represents the ideal influence matrix, obtained when no errors are present in the perturbation data, and the right panel represents a realization of the influence graph with both false positive and false negative edges. In the first step of RIPE, causal orderings are determined based on the graph in the right panel of Figure~1D. 
An example of such an ordering is $o = (g_2 \prec g_1 \prec g_3 \prec g_4 \prec g_5)$. Note that
in this case many orderings exist, due to the presence of cycles in the influence graph. In the second step of the RIPE algorithm, a penalized regression problem is solved for each node, where the set of predictors are the set of parents of the node in the right panel of Figure~1D, 
consistent with the given ordering. In particular, the following regression problems are solved for $o$ (here $X_i \sim X_j + X_k$ denotes regression of $X_j$ on $X_i$ and $X_k$, {\em ignoring} for ease of presentation the corresponding penalty term):
\begin{equation*}
    X_1 \sim X_2,  \;  
    X_3 \sim X_1 + X_2,  \; 
    X_4 \sim X_1 + X_2, \; 
    X_5 \sim X_2 + X_3 + X_4
\end{equation*}
Using the results of these regressions, the value of the penalized negative log-likelihood function is then determined for each of the estimated graphs.
Based on these values, the ``best'' networks are then used to construct the consensus graph, as described next in the section discussing the third step of the RIPE algorithm.

Given the set of orderings $\mathcal{O}$ of cardinality $M\equiv|\mathcal{O}|$, one needs to solve $M$ separate penalized regression problems, and store their
corresponding penalized negative log-likelihood values, where the computational cost of each of these problems is $O(n p^2)$. However, this step is fully
parallelizable and using $\nu$ processors the complexity reduces to $O(\frac{M}{\nu}n p^2)$.

\subsection*{Graph averaging: a consensus regulatory network}\label{model_4}
As mentioned above, the estimated influence matrix $\mathcal{P}$ from perturbation screens results in multiple orderings $o$, either due
to feedback regulatory mechanisms or noisy measurements.
In such cases, the estimate of the adjacency matrix of the graph can be obtained from those corresponding to the set of orderings achieving the smallest penalized negative log-likelihood values. Therefore, the final step of the RIPE algorithm includes a model averaging procedure that combines the estimated DAGs from multiple orderings to construct a {\em cyclic consensus network}. This is illustrated with a small cyclic subnetwork in Figure~S2. In this example, the true network includes a number of cycles, and the estimate from the RIPE algorithm correctly identifies some of these cycles (also see Text~S1).

Let $L_q$ denote the lower $q$th quantile of the penalized negative log-likelihood values, denoted by $\ell$,
and $Q = \{o \in \mathcal{O}: \ell(o) \le L_q \}$ be the set of orderings with the lowest $100 q\%$ penalized negative log-likelihood values. The RIPE estimate of the adjacency matrix is then defined as the consensus DAG:
\begin{equation} \label{RIPEest}
       \hat{A}^{c}_{i,j} = \frac{1}{|Q|} \sum_{k \in Q}{ \mathbf{1}_{ |A^{k}_{i,j}| > 0 } }
\end{equation}
Here, $A^{k}$ is the DAG estimate from the $k$-th ordering, $\hat{A}^{c}$ denotes the \emph{confidence} of each edge in the final estimate, and
$ \mathbf{1}_{ \Omega }$ is the indicator function which equals $1$ when the condition $\Omega$ holds, and $0$ otherwise.

Consequently, the estimate of the edge set of the graph is defined as $\hat{E} = \{ (i,j): \hat{A}^{c}_{i,j} \ge \tau \}$, where $\tau$ is the threshold for including an edge. The value of $\tau$ determines the desired level of confidence, and can be chosen by the user depending on the application of interest. Nevertheless, the above formulation provides a flexible estimation framework, in which $\tau$ is considered a tuning parameter or can be set based on prior analyses (see the section Choice of Parameters and Properties of the Algorithm below and the Results section for more details about the choice of $\tau$).

The RIPE algorithm also produces an estimate of the \emph{sign} of each edge, as well as the \emph{magnitude} of the effect, defined by the matrices $\hat{A}^{s}$ and $\hat{A}^{v}$ below:
\begin{equation} \label{RIPEest}
       \hat{A}^{s}_{i,j} = \sgn\Big( \sum_{k \in Q}{ \sgn( A^{k}_{i,j} ) } \Big) \qquad
       \hat{A}^{v}_{i,j} =  \frac{1}{|Q|}\sum_{k \in Q}{ | A^{k}_{i,j} | }
\end{equation}
where the $\sgn(0) \equiv 0$.

\subsection*{Choice of Parameters and Properties of the Algorithm}\label{model_5}
Similarly to any other learning algorithm, the performance of the RIPE algorithm depends on the choices of tuning parameters. There are three tuning parameter that need to be determined: (i) the penalty coefficient $\lambda$, (ii) the likelihood quantile $q$, and (iii) the confidence threshold $\tau$. Next, we discuss strategies for choosing these parameters.

In penalized regression settings, $\lambda$ determines the weight of the penalty term in the optimization problem, with larger values
of $\lambda$ resulting in more sparse estimates. In \cite{shojaie2010penalized}, the following error-based choice of $\lambda$ was
proposed for the $i$-th regression in ~\eqref{DAGest_lasso},
\begin{equation}\label{eqn:lambda}
    \lambda_i(\alpha) = 2 n^{-1/2} Z^*_{\frac{\alpha}{2p(i-1)}}.
\end{equation}
Here, $Z^\ast_{x}$ is the $x$-quantile of the standard normal distribution and it can be shown that this choice of penalty controls the probability of falsely joining two unconnected ancestral components at level $\alpha$. Interestingly, numerical studies in \cite{shojaie2010penalized} show that the result of the analysis is not sensitive to the choice of $\alpha$, and values of $\alpha \in (0.1, 0.5)$ result in comparable estimates.

Even though the choice of $\lambda$ in equation~\eqref{eqn:lambda} controls the probability of false positives, it results in over-sparse estimates. Numerical studies in \cite{shojaie2011thresholding} strongly
suggest the smaller value $0.6 \, \lambda_i(\alpha)$, which is used in the numerical studies in this paper.

Unlike the choice of $\lambda$, our numerical studies indicate that the RIPE algorithm is not sensitive to the choices of $q$ and $\tau$, and
a wide range of values can be used for these parameters (see Results and Figures~S3-S5). It is worth pointing out that since the value of $q$
determines the proportion of highest likelihoods used in constructing the consensus graph, when the perturbation data gives a reliable estimate
of the influence matrix $\mathcal{P}$, multiple orderings from the first step of RIPE would result in true feedback cycles in regulatory networks, and therefore,
the differences in corresponding likelihood values are mostly due to the inherent noise in expression data. As a result, in the
ideal case with no errors in the influence graph,
a value of $q = 1$ would produce the ``best'' estimate. However, in practice, to avoid inferior estimates, outlier
values of the likelihood should not be incorporated in the estimate. Our numerical analyses show that values of $q \in (0.1, 0.9)$ would result in comparable estimates.

Finally, the choice of $\tau$ determines the confidence of edges in the estimated consensus network: large values of $\tau$ result in edges that are more
consistently present in all estimated graphs, while small values allow for less frequently present edges to be included in the final estimate.
As with $\lambda$, our numerical studies indicate that large values of $\tau$ result in over-sparse estimates, and we recommend
values of $\tau \in (0.05, 0.35)$ (see the Results section and Figures~S3-S5). All estimates in the paper were obtained by
setting $q = 0.1$ and $\tau = 0.25$, to achieve a balance between the standard performance measures of
Precision and Recall. 

To complete our discussion of the RIPE algorithm, it is worth noting that an accurate influence matrix $\mathcal{P}$ including binary information from single gene knockouts contains sufficient information for obtaining the causal orderings of the underlying regulatory network, but {\em not} its structure (see Lemma~1 in the Supporting Information (Text S1) for a formal statement and proof of this result). In the RIPE algorithm, abundantly available steady-state expression data are used to compensate for this lack of information. Further, we
establish the asymptotic consistency of the network structure estimated using the penalized likelihood method (see Lemma~2 in the Supporting Information, Text S1).

\begin{figure}[!h]
 \begin{center}
\includegraphics[width=6.3in]{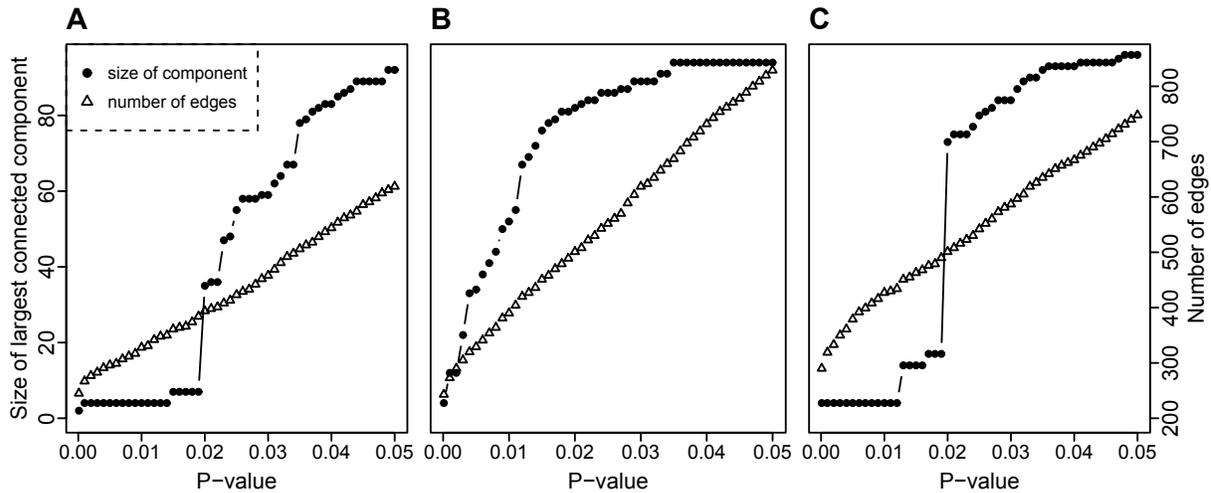}
\end{center}
\caption{
{\bf Influence graph characteristics versus $p$-value for DREAM4.} Number of edges (open triangles) in the influence graph and the size of the largest connected component (dots) versus cut-off $p$-value for differential expression.
The data is based on five replications of the knockout and wild-type experiments for (A) 100-node network 1, (B) 100-node network 3, and (C) 100-node network 5 in the DREAM4 challenge.}
\label{fig:dreampvalKO}
\end{figure}
\section*{Results}\label{result}
\subsection*{Preliminaries}\label{prelim}
\subsubsection*{Determining the Influence Matrix}
In Section Estimating the Influence Matrix from Perturbation Data under Methods we describe how to estimate an influence matrix $\mathcal{P}$ from perturbation experiments using differential expression analysis. The analysis produces $p$-values for each entry in the matrix, and by choosing a specific cutoff $p^\ast$, we get an estimate of $\mathcal{P}$; note that a lower $p^\ast$ gives rise to a sparser matrix. To select a reasonable cutoff, the number of edges in the influence graph for different p-values are plotted together with the size of the largest connected component, see Figure~2. 

If the network is modularized, i.e. to some extent consists of nodes grouped together as illustrated in Figure~1B, 
we would consider a drop in the size of the largest connected component indicative of a good choice of $p$-value. If only a few edges make the difference between a large and a small connected component, we have most likely found the $p$-value for which the ``noise edges'' have been removed to reveal the modularized structure. If such a drop in the size of the large component is missing, possible reasons can be that the underlying network does not exhibit a modularized structure, or that the signal strength of the experimental data is rather low to clearly reveal it. However, a plot of the type given in Figure~2 
can still be useful in such a setting, since we can choose a $p$-value based on the size of the connected component. An overly large connected component is not very realistic from a biological perspective, in addition to being computational demanding since it produces many potential orderings.

\subsubsection*{Performance Evaluation Criteria}
Several evaluation criteria have been proposed for assessing the performance of different network reconstruction methods. However, the choice of
criteria depends on the information available on the ``true'' regulatory network: in synthetic, or in silico examples, the gold standard is known,
and Precision, Recall and $F_1$ measures are often used as standard metrics (with $F_1$ being the
harmonic mean of the other two metrics). On the other hand, in real data applications, the gold standard is often unknown and
the appropriate performance criteria should be determined based on the information at hand. In the section below, titled
Regulatory Network in Yeast, the known protein-DNA interactions for yeast present in the BIOGRID database (release 3.1.74) \cite{stark2006biogrid} are
used as gold standards. To account for the fact that the BIOGRID database may not include all regulatory interactions, we carried out an experiment similar to
the approach implemented in \cite{maathuis2010predicting}. More specifically, to assess the significance of the number of true positives for each of the
estimates, 10,000 Erd\"{o}s-R\'{e}nyi random graphs with the same number of edges as each of the estimates were
generated (using the R-package \texttt{igraph} \cite{igraph}), and the number of true positives in randomly generated
graphs was used to approximate the $p$-value for significance of the number of true positives observed. The resulting
$p$-value determines the likelihood of observing a given number of true positives in randomly generated graphs, and can be used directly for performance evaluation.

\subsection*{DREAM4 Challenge}\label{dream}
To illustrate the performance of the RIPE algorithm, we first evaluate it by using networks from the DREAM4 in-silico network challenge, which is part of a series of challenges \cite{prill2010dream3}, aiming at inferring gene regulatory networks from simulated data. The gold standard networks in DREAM4 were generated by extracting modules from two source networks:
yeast and  \emph{Escherichia coli}. The modules were extracted so as to preserve the structural properties of the underlying network, such as degree distribution and network motifs (statistically overrepresented circuit elements) \cite{marbach2009gen}. The underlying dynamics in the models include mRNA and protein dynamics simulated by stochastic differential equations with both experimental and internal noise added (an extension to the method in \cite{marbach2010rev}).

The simulated data sets available within DREAM4 consist of observations from the unperturbed network (\emph{wild-type}), perturbation experiments in which all genes are knocked-out one-by-one (\emph{knockouts}), perturbation experiments in which the activity of all genes are lowered one by one with a factor of two (\emph{knockdowns}), as well as small perturbation of all genes simultaneously (\emph{multifactorial}) and finally, time series.

Three of the five 100-node networks were selected for analysis; network 1 (\emph{Net1}), network 3 (\emph{Net3}) and network 5 (\emph{Net5}). These particular choices were made based on the differences in topologies, as well as on how well the network structures were predicted in terms of AUC (Area Under the ROC Curve) in the in-silico challenge. Specifically, several competing research teams submitted predictions of the network structures for the DREAM4 challenge. \emph{Net1} was best predicted overall, while the structure of \emph{Net5} was the most difficult to deduce. We chose to also analyze \emph{Net3} since Networks 2-4 were predicted with comparable accuracy, with intermediate AUC values compared to \emph{Net1} and \emph{Net5}. The varying prediction accuracies can be explained by the differences exhibited by the various network structures; for example, \emph{Net1} has a layered structure, with many nodes functioning as either parents or children in the graph, while \emph{Net5} has a more complex topology with several short cycles and nodes with multiple parents.

The DREAM4 challenge only includes one replicate of each simulated experiment, and in order to assess the noise levels in the data, we simulated five replicates of each of the wild-type, knockdown, and knockout experiments, as well as one multifactorial data set for the networks of interest by using GeneNetWeaver 3.0. The DREAM4 default settings were used, excluding standardization of the simulated data.

Two methods tied for first place in predicting the topology for the 100-node networks in DREAM4; 
a $t$-test based method \cite{greenfield2010infer} and a
method based on confidence matrices, pruned by down-ranking of feed-forward loops (FFLDR) \cite{pinna2010}. As the two methods performed comparably,
we chose to compare our method to FFLDR, for which the authors kindly provided their code. In \cite{greenfield2010infer}, the $t$-test method is also
combined with time-series data using ordinary differential equations to model the temporal changes in the gene expression (known as Inferelator).
However, as this method was not submitted in the challenge of network topology prediction, and in addition utilizes time series data, we have not
included it in the comparisons.

The influence matrix $\mathcal{P}$ was estimated by comparing the expression levels from perturbation experiments (five replicates for each knockout/knockdown) to the corresponding levels in the unperturbed network
(five replicates of the wild-type data), as described in the Preliminaries section above (see Figure~2). 
The $p$-value 0.019 was selected for both \emph{Net1} and \emph{Net5}, while 0.003 was chosen for \emph{Net3}. For the knockdown data, the same method was used to select the cutoff p-values as for the knockout data (see Figure~S6). 
For FFLDR, the optimal performance for large-scale networks is obtained by directly comparing the expression levels in knockout experiments, and hence, the five replicates of the knockout data (not using the wild-type data for comparison) were averaged, and this signal was used as input to the method.

Table~\ref{tbl:DREAM_sim} summarizes the performances of the PC-algorithm (PCALG, as implemented in the R-package \texttt{pcalg}), the Nested Effects Model
(NEM, as implemented in the R-package \texttt{nem}), as well as the FFLDR and RIPE (based on 10,000 MC-DFS orderings) algorithms in reconstructing \emph{Net1}, \emph{Net3}, and \emph{Net5}. Since the DREAM4 challenge does not include suitable steady-state expression data, to employ PCALG and RIPE, $n=100$ independent samples from a Gaussian distribution with mean $0$ and variance $1$ were generated from each of the networks. The strengths of associations among genes were chosen as uniform random numbers in the range $\pm(0.2, 0.8)$ and to handle the effect of cycles in the regulatory network, the magnitudes of strengths were then normalized according to the method described in \cite{shojaie2010network}.

\begin{table}[t]
\caption{{\bf Reconstruction results for DREAM4 networks with simulated steady state data.}}
\centering
{\footnotesize
\begin{tabular}{|l|ccc|ccc|ccc|ccc|ccc|ccc|} \hline
\,	& \multicolumn{3}{|c|}{Net1, KO} & \multicolumn{3}{|c|}{\emph{Net1}, KD} &
\multicolumn{3}{|c|}{\emph{Net3}, KO} & \multicolumn{3}{|c|}{\emph{Net3}, KD} &
\multicolumn{3}{|c|}{\emph{Net5}, KO} & \multicolumn{3}{|c|}{\emph{Net5}, KD} \\ \cline{2-19}
Method & P & R & F$_1$ & P & R & F$_1$ & P & R & F$_1$ & P & R & F$_1$ & P & R & F$_1$ & P & R & F$_1$ \\ \hline
ARACNE & 17 & 41 & 24 & 17 & 41 & 24 & 10 & 48 & 16 & 10 & 48 & 16 & 16 & 51 & 25 & 16 & 51 & 25 \\
PCALG &	48 & 26 & 34 & 40 & 23 & 29 & 48 & 25 & 33 & 44 & 25 & 31 & 40 & 20 & 27 & 43 & 22 & 29 \\
NEM	  & 28 & 34 & 31 &  6 &  6 &  6 & 14 & 9  & 11 &  4 &  4 &  4 &  9 & 17 & 12 &  7 & 10 & 08 \\
FFLDR &	92 & 57 & 70 & 86 & 46 & 60 & 66 & 44 & 53 & 60 & 28 & 38 & 44 & 35 & 39 & 49 & 26 & 34 \\
RIPE  &	80 & 71 & 75 & 74 & 46 & 56 & 65 & 49 & 56 & 61 & 30 & 40 & 59 & 47 & 52 & 56 & 34 & 42 \\ 
\hline
\end{tabular}
}
\begin{flushleft}
Performance measures, in percentages, for methods of reconstruction of DREAM4 \emph{Net1}, \emph{Net3}, and \emph{Net5}, using both knockout (KO) and knockdown (KD) data. Steady state expression data is generated from structural equation models based on the true graph.
\end{flushleft}
\label{tbl:DREAM_sim}
\end{table}
\begin{table}[t]
\centering
\caption{{\bf Reconstruction results for DREAM4 networks with multifactorial data.}}
{\footnotesize
\begin{tabular}{|l|ccc|ccc|ccc|ccc|ccc|ccc|ccc|ccc|} \hline
\,	& \multicolumn{3}{|c|}{\emph{Net1}, KO} & \multicolumn{3}{|c|}{\emph{Net1}, KD} &
\multicolumn{3}{|c|}{\emph{Net3}, KO} & \multicolumn{3}{|c|}{\emph{Net3}, KD} &
\multicolumn{3}{|c|}{\emph{Net5}, KO} & \multicolumn{3}{|c|}{\emph{Net5}, KD} \\ \cline{2-19}
Method & P & R & F$_1$ & P & R & F$_1$ & P & R & F$_1$ & P & R & F$_1$ & P & R & F$_1$ & P & R & F$_1$ \\ \hline
ARACNE & 25 & 9 &  13 & 25 & 9 & 13 &  28 & 19 & 23 & 28 & 19 & 23 & 16 & 14 & 15 & 16 & 14 & 15 \\
PCALG  & 24 & 09 & 13 & 24 & 09 & 13 & 39 & 15 & 22 & 39 & 15 & 22 & 21 & 08 & 12 & 21 & 08 & 12 \\
NEM	   & 28 & 34 & 31 &  6 &  6 &  6 & 14 & 9  & 11 & 4  & 4  & 4  &  9 & 17 & 12 &  7 & 10 &  8 \\
FFLDR  & 92 & 57 & 70 & 86 & 46 & 60 & 66 & 44 & 53 & 60 & 28 & 38 & 44 & 35 & 39 & 49 & 26 & 34 \\
RIPE   & 74 & 38 & 50 & 80 & 30 & 43 & 71 & 31 & 43 & 68 & 23 & 34 & 54 & 32 & 40 & 52 & 24 & 33 \\
\hline
\end{tabular}
}
\begin{flushleft}
Performance measures, in percentages, for methods of reconstruction of DREAM4 \emph{Net1}, \emph{Net3}, and \emph{Net5}, using both knockout (KO) and knockdown (KD) data. Multifactorial data from DREAM4 challenge is used as steady-state expression data.
\end{flushleft}
\label{tbl:DREAM_multfac}
\end{table}
As a benchmark, we also evaluated the performance of the ARACNE procedure \cite{margolin2006aracne} on the simulated data. It is important to note that ARACNE estimates \textit{undirected} networks. Thus, the estimates from ARACNE cannot be directly compared with the other methods mentioned above. Nonetheless, we applied ARACNE with a number of $p$-value cutoffs and found that the Bonferroni adjusted $p$-value cutoff of $1.010 \times 10^{-5}$ offers the best performance compared to the true network.

It can be seen from Table~\ref{tbl:DREAM_sim} that RIPE outperforms the competing methods (see $F_1$ measure), with the exception of knockdown data for \emph{Net1} where FFLDR exhibits a slight edge. The differences in performance are more pronounced for the more complex \emph{Net5}.
As mentioned above, \emph{Net3} and \emph{Net5} pose more challenging reconstruction problems compared to \emph{Net1}; hence, all methods exhibit inferior performances in the reconstruction task. On the other hand, while knockout data represent ideal perturbation experiments, knockdown data correspond to less accurate ones. Therefore, the performances of NEM and FFLDR, that only employ perturbation screens, as well as RIPE, deteriorate in the case of knockdown data, whereas the performance of PCALG and ARACNE are not affected by the change in the perturbation data, as it uses as input only steady state expression data.

To further evaluate the performance of ARACNE, PCALG and RIPE, we also applied these methods to the multifactorial data. As described earlier, the multifactorial data set is obtained from non-i.i.d observations, which violate the underlying assumption of both PCALG and RIPE algorithms. In addition, this data set does not correspond to a steady-state setting.
Interestingly, the results in Table~\ref{tbl:DREAM_multfac} indicate that the performance of all three competitors deteriorates by roughly similar 
factors, which can be attributed to the lower quality of this multifactorial data set. On the other hand, the results show that even when the steady-state data violate the underlying assumptions of the model, the performance of the RIPE algorithm is comparable to that of FFLDR, particularly for the more complex structure of \emph{Net5}.

These results strongly indicate that combining perturbation screens with steady state expression data are beneficial to the regulatory network reconstruction problem, especially in settings where the perturbation data are rather noisy and the steady state data exhibit good quality. To better address this issue and obtain a deeper insight into the effect of noise on the perturbation data we undertake a number of experiments based on synthetic data.

\subsection*{Experiments with Synthetic Data}\label{sim}
To assess the influence of the inputs and steps required by the various algorithms, we examine a number of settings both in small and large scale networks.

\subsubsection*{Small Directed Acyclic Graph}\label{sim1}
We start our discussion on synthetic data with the toy example illustrated in Figure~S7. 
Specifically, we employ a randomly generated DAG of size $p=20$ corresponding to the true regulatory network.
To emulate possible regulatory mechanisms, the generated DAG includes a number of ``hub'' genes, as well as two genes that are not regulated by any other gene.

To obtain independent expression data ($\mathcal{X}$) consistent with the underlying DAG, an association weight of $\rho = 0.8$ is assigned to all the edges in the graph and the available functions in the R-package \texttt{pcalg} are used to generate $n = 50$ independent samples of Gaussian random variables with mean 0 and variance 1, according to structural equation models that incorporate the influence of nodes of the graph on each other (see \cite{kalisch2010causal} for more details).

\begin{table}[t]
\centering
\caption{{\bf Reconstruction results for synthetic networks subject to error in the perturbation data.}}
{\footnotesize
\begin{tabular}{|l|ccc|ccc|ccc|ccc|} \hline
\,	& \multicolumn{3}{|c|}{$\mathcal{P}_0$} & \multicolumn{3}{|c|}{$\mathcal{P}_1$} & \multicolumn{3}{|c|}{$\mathcal{P}_2$} & \multicolumn{3}{|c|}{$\mathcal{P}_3$}  \\ \cline{2-13}
Method	&	P	&	R	&	$F_1$	&	P	&	R	&	$F_1$	&	P	&	R	&	$F_1$	&	P	&	R	&	$F_1$	\\ \hline
PCALG	& 54(8) & 52(10) & 53(8) & 54(8) & 54(8) & 52(10) &  53(8) & 52(10) & 53(8) & 54(8) & 52(10) & 53(8)	\\
NEM	    & 50	&  33 & 40	&	55	&	33	&	39	&	36	&	33	&	41	&	21	&	28	&	24	\\
FFLDR	&	100	&	67	&	79	&	40	&	92	&	67	&	77	&	33	&	36	&	33	&	28	&	30	\\
RIPE	& 95(4) & 99(2) & 97(2) & 79(4) & 90(4) & 94(1) & 92(3) & 100(1) & 88(3) & 61(5) & 84(6) & 71(5) \\
\hline
\end{tabular}
}
\begin{flushleft}
Average performances measures, in percentages, for methods of network reconstruction in the synthetic networks. Numbers in parentheses show standard deviations for methods based on simulated steady state data (PCALG and RIPE). $\mathcal{P}_0$ indicates the ideal influence graph and $\mathcal{P}_1$ to $\mathcal{P}_3$ represent noisy versions of the influence graph (see Figure~S8).
\end{flushleft}
\label{tbl:simperf}
\end{table}
The influence matrix $\mathcal{P}$ corresponding to the perturbation data is generated as follows: given the adjacency matrix of a DAG $A$, it is shown in \cite{ShojMich:09} that $\mathcal{P}=\lceil (I-A)^{-1} \rceil$, where $I$ denotes the identity matrix and $\lceil \cdot \rceil$ the ceiling operator. We denote by $\mathcal{P}_0$ the ground truth influence matrix corresponding to the generated DAG.
However, as previously discussed, in practice the influence matrix is extracted from gene expression data obtained from the perturbation experiments and hence is inherently noisy. Depicted in Figure~S8 
is the matrix $\mathcal{P}_0$  (left-most image) and three variants of that matrix that we examine; in the first one ($\mathcal{P}_1$) the direction of a small proportion of edges reversed (second image from the left) and in the second one ($\mathcal{P}_2$) edges are added (second image from the right); the third matrix ($\mathcal{P}_3$) includes both reversed edges, as well as an addition of extra edges (right-most image).

In this case, the small size of the network together with the relatively small amount of noise introduced in the influence matrix allows us to obtain {\em all} possible orderings; specifically, $\mathcal{P}_3$ which contains the most number of orderings, has 12 strongly connected components of small size with a total of 3926 possible orderings, and hence can be easily handled with exhaustive search. On the other hand, the total number of orderings for $\mathcal{P}_0$, $\mathcal{P}_1$ and $\mathcal{P}_2$ are 1, 1, and 2 respectively. However, we emphasize that obtaining the causal orderings of
large strong components (i.e., larger than 15 nodes) using the exhaustive  method is computationally expensive, and the MC-DFS heuristic becomes the only practical choice.

\begin{figure}[!t]
 \begin{center}
\includegraphics[width=15cm]{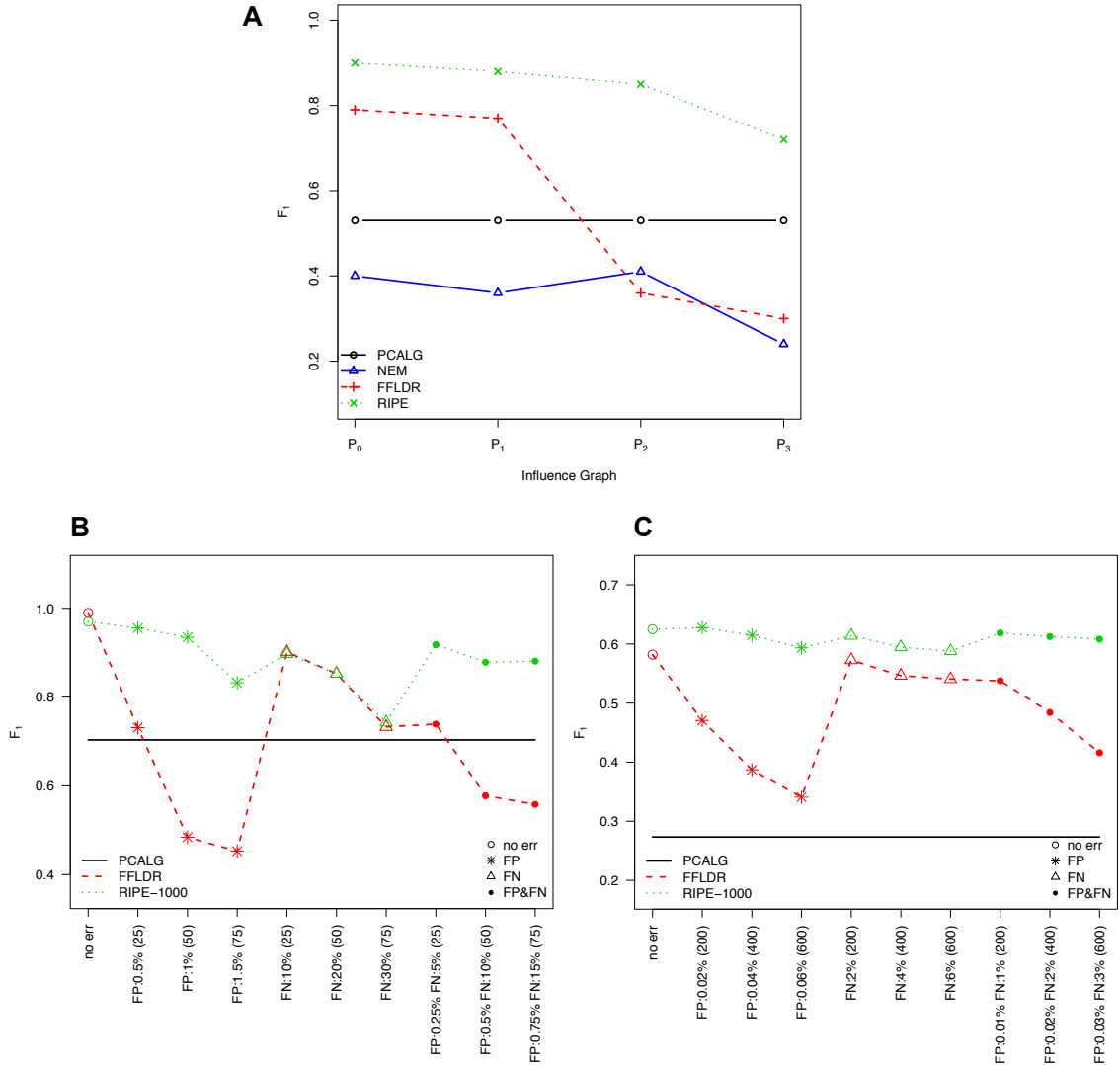}
\end{center}
\caption{{\bf Performance of RIPE and competing methods on the reconstruction of synthetic networks.}
(A) Average $F_1$ measures for reconstruction using NEM, PCALG, FFLDR and RIPE on a network of size $p=20$. $\mathcal{P}_0$ corresponds to the ideal influence graph and $\mathcal{P}_1$ to $\mathcal{P}_3$ represent increasing levels of noise in perturbation data (see also Figure~S8).
(B)-(C) Average $F_1$ measures for reconstruction of synthetic regulatory networks using PCALG, FFLDR and RIPE for different levels of false positive and negative noise in perturbation data. Numbers in parentheses indicate the expected number of false edges in each case. The true graph is an acyclic graph (DAG) of size $p=100$ in (B) and a cyclic graph of size $p=1000$ in (C).}
\label{fig:simperf}
\end{figure}
The results of applying the four methods under consideration to each of the input data sets are given in Table~\ref{tbl:simperf}. Note that NEM and FFLDR use only the four influence matrices $\mathcal{P}_0-\mathcal{P}_3$ as input, while the PCALG only uses the steady state gene expression data $\mathcal{X}$; finally,  RIPE uses both of them. The numbers in parentheses correspond to the standard deviation of the metrics for PCALG and RIPE obtained from $100$ replications of the steady state gene expression data $\mathcal{X}$. The numbers reported for RIPE, are obtained by considering all possible orderings generated using the exhaustive DFS algorithm for $\mathcal{P}_0-\mathcal{P}_3$, and for $\mathcal{P}_3$ the consensus graph was obtained by setting $q = 0.1$ and $\tau = 0.25$.
A graphical summary of $F_1$ measures over different versions of the influence matrix is given in Figure~3A. 
It is worth noting that as discussed above, the choices of $q$ and confidence threshold $\tau$ do not critically affect the performance of RIPE and values of $q \in (0.1, 0.9)$ and $\tau \in (0.05, 0.6)$ yield comparable results (see Figures~S3-S5). 

It can be seen that by combining perturbation screens and steady state gene expression data, RIPE clearly outperforms the FFLDR, NEM and PCALG methods that use a single source of data. Note that PCALG only uses the $\mathcal{X}$ data, thus the identical entries in the table. It is worth noting that although the performances of NEM, FFLDR and RIPE are affected by the increasing level of noise in the perturbation data (as expected), RIPE can compensate for this loss of accuracy by incorporating the additional information from the steady-state data. Obviously, all these effects would be magnified if a large number of spurious edges are added or a large number of true edges are deleted, thus introducing a significant amount of noise in the influence matrix $\mathcal{P}$.
Interestingly, our results indicate that PCALG has a slight edge over NEM (both in case of DREAM data and the synthetic network). This may be due to the specific structure of the DAG under consideration (experiments show that NEM, which uses triplets of nodes to determine the order of the edges, often performs better in chain-type graphs, for example in DREAM4 \emph{Net1}).
The inferior performance of NEM in this setting can also be attributed to the fact that NEM, as originally proposed, performs well settings where the number of perturbation experiments is significantly smaller than the number of affected genes, which is not the case in our numerical experiments.
On the other hand, RIPE takes a global view by constructing causal orderings, in addition to independently evaluating them with steady state gene expression data. Of particular interest is the significant deterioration in the performance of FFLDR in the cases of $\mathcal{P}_2$ and $\mathcal{P}_3$ that indicates a vulnerability of the method in the presence of noise in its input data.

\begin{table}[t]
\centering
\caption{{\bf Impact of increasing number of orderings used in the RIPE algorithm.}}
{\footnotesize
\begin{tabular}{|l|ccc|} \hline
Number of Orderings	&	P	&	R	&	$F_1$	\\ \hline
100	    &	60	&	81	&	69	\\
200	    &	61	&	82	&	70	\\
1000	&	61	&	82	&	70	\\
2000	&	60	&	83	&	70	\\
3926 (all)	&	61	&	84	&	71	\\ \hline
\end{tabular}
}
\begin{flushleft}
Average performance measures, in percentages, for RIPE in the synthetic network $\mathcal{P}_3$.
\end{flushleft}
\label{tbl:ordcomp}
\end{table}
Finally, to assess the effect of approximation used in MC-DFS, in comparison to having the universe of orderings, we estimated the regulatory network from $\mathcal{P}_3$ with different number of orderings. The average values of the Precision, Recall and $F_1$ measures over 100 replications for different number of orderings considered are given in Table~\ref{tbl:ordcomp}. All estimates were obtained by using $q=0.1$ and $\tau = 0.25$ for estimating the consensus graph. It can be seen that, in comparison to the estimate obtained by evaluating all 3926 ordering using exhaustive DFS, the one obtained by considering a random subset of a small number of orderings provides comparable results. In particular, aside from a slight increase in recall, as the number of orderings increases, the performance of the method with only 200 orderings obtained using MC-DFS  is comparable to utilizing the universe of orderings generated from an exhaustive search with DFS.
This result suggests that the MC-DFS algorithm represents a viable alternative for settings involving a large size strongly connected component, and as a result, the RIPE algorithm is not highly sensitive to the number of orderings used to reconstruct the network.

\subsubsection*{Effect of False Positive and Negative Errors in Perturbation Data}\label{sim2}
To analyze the effect of false positive and negative errors in the perturbation data on the performance of the RIPE algorithm, a randomly generated DAG of size $p=100$ was used as the true regulatory network.
The influence matrix $\mathcal{P}$ corresponding to the true perturbation data, and independent expression data of size $n = 100$ consistent with the underlying DAG were generated according to the methods explained in the previous section.

To emulate the effect of false positive and negative errors, three different scenarios were considered: false positive errors ($FP$), in the form of edges not present in the true influence graph, false negative edges ($FN$), in the form of missing edges compared to the true graph, and both false positive and false negative edges ($FP+FN$). Additionally, to assess the effect of increasing noise levels, 3 levels of noise in each of the above scenarios were considered. To determine the appropriate noise levels, the number of edges in $\mathcal{P}_0$ (198) were used as calibration, and the proportion of false positive and false negative edges were adjusted so that each randomly perturbed influence matrix included the same number of expected false edges. The results were 9 noisy influence graphs under each of $FP$, $FN$ and $FP+FN$ settings with roughly 25, 50 and 75 false edges.

Considering the inefficiency of NEM for network estimation in high dimensional settings especially when the number of effect genes is small, we focus on the performance of FFLDR and RIPE, using PCALG as a benchmark. The results for RIPE were obtained based on 1000 randomly generated orderings for the cases where the influence graphs were generated with false positives (since the true graph is a DAG, the case of $FN$ amounts to a single ordering).

The performance of the above methods in each of the input data sets are given in Table~\ref{tbl:errcomp100}, where the results for RIPE and PCALG correspond to averages over $50$ independent draws of the steady state data. Figure~3B 
summarizes the values of $F_1$ measures for different methods.
These results reveal a number of interesting aspects of FFLDR and RIPE algorithms. First, as expected, increasing levels of false negatives impact the performance of RIPE, while FFLDR could be severely impacted by high false positive rates. Secondly, while the worst-case performance of RIPE (for the case of $FN$) matches that of FFLDR, the proposed data integration framework can result in significant gain in network estimation accuracy in other scenarios. While high quality perturbation screens greatly improve the performance of both, by combining steady state data and perturbation screens, the RIPE algorithm could compensate for the inaccuracy of the perturbation data, whereas estimation based on perturbation data alone can result in FFLDR estimates that are inferior to those of PCALG. Finally, Figure~S9 
shows the improvement with increasing number of orderings in $F_1$, $P$, and $R$ for inference using the influence graph with highest level of false positives ($1.5\%$). It can be seen that although the three measures moderately improve with higher number of orderings, a small number of MC-DFS orderings are sufficient for acceptable performance of the RIPE algorithm.
Note that due to the acyclicity of the underlying graph, different levels of false negative errors correspond to a single ordering for the influence graph, and hence a similar comparison for the case of false negatives is not relevant.

\subsubsection*{Large Cyclic Graph ($p=1000$)}\label{sim2}
Our final numerical experiment with synthetic data compares the performance of RIPE with those of PCALG and FFLDR in reconstructing large cyclic graphs in the presence of both false positive and negative noise in the perturbation data. The setting of this simulation is similar to that of the previous section, with the main difference being the size of the graph and presence of cycles (feedback loops) in the true graph. In particular, a random cyclic graph with $p = 1000$ nodes and $\sim 2p = 1984$ edges was generated, and $n = 500$ samples were generated from zero mean, unit variance Gaussian random variables, as steady state expression levels from the true network.

As before, three different scenarios were considered: false positive errors ($FP$), false negative edges ($FN$), and both false positive and false negative edges ($FP+FN$) at 3 noise levels in each of the above scenarios. The noise levels were set up so that approximately 200, 400, and 600 erroneous edges were included in each of $FP$, $FN$, and $FP+FN$ settings.

The performance of the above methods for each of the input settings is given in Table~\ref{tbl:errcomp1000}, where the results for RIPE (with 1000 random orderings) and PCALG are averages over $5$ independent drawings of the steady state data.
Figure~3C 
summarizes the values of $F_1$ measures for different methods. These results confirm the findings of the previous section. In particular, RIPE outperforms the other two algorithms in all the simulated settings, and the difference between the performances of RIPE and FFLDR is magnified as more false positive edges are added to the perturbation graph. Finally, as expected, the PCALG does not compare favorably with the other two methods in the setting of cyclic graphs.

\begin{table}[t]
\centering
\caption{{\bf Impact of increasing false positive and negative errors in perturbation data on estimation of acyclic graphs.}}
{\footnotesize
\begin{tabular}{r|l|ccc|ccc|ccc|}
\multicolumn{1}{c}{\,} & \multicolumn{1}{c}{\,} & \multicolumn{3}{c}{$F_1$} & \multicolumn{3}{c}{$P$} & \multicolumn{3}{c}{$R$} \\ \cline{3-11}
\multicolumn{1}{c}{\,} & \multicolumn{1}{c|}{\,} & PCALG & FFLDR & RIPE & PCALG & FFLDR & RIPE & PCALG & FFLDR & RIPE \\ \cline{2-11} \cline{2-11}
\,     & NO ERR          & 70(3) & 99 & 97(1) & 68(3) &100 & 97(2) & 73(3) & 98 & 97(1) \\ \cline{2-11}
\,     & $0.50\%$        & 70(3) & 73 & 96(1) & 68(3) & 65 & 95(2) & 73(3) & 84 & 97(1) \\
$FP$   & $1\%$           & 70(3) & 48 & 94(1) & 68(3) & 40 & 91(2) & 73(3) & 62 & 96(1) \\
\,	   & $1.50\%$        & 70(3) & 45 & 83(1) & 68(3) & 30 & 77(2) & 73(3) & 97 & 91(1) \\ \cline{2-11}
\,	   & $10\%$          & 70(3) & 90 & 90(1) & 68(3) & 93 & 92(2) & 73(3) & 88 & 87(1) \\
$FN$   & $20\%$          & 70(3) & 85 & 85(1) & 68(3) & 86 & 88(1) & 73(3) & 85 & 83(1) \\
\,	   & $30\%$          & 70(3) & 73 & 74(1) & 68(3) & 76 & 80(2) & 73(3) & 71 & 69(1) \\ \cline{2-11}
\,	   & $0.25\%$, $5\%$ & 70(3) & 74 & 92(1) & 68(3) & 65 & 92(2) & 73(3) & 86 & 93(2) \\
$FP+FN$& $0.5\%$, $10\%$ & 70(3) & 58 & 88(1) & 68(3) & 52 & 88(2) & 73(3) & 66 & 87(1) \\
\,     & $0.75\%$, $15\%$& 70(3) & 56 & 88(1) & 68(3) & 48 & 89(2) & 73(3) & 68 & 87(1) \\ \cline{2-11} \cline{2-11}
\end{tabular}
}
\begin{flushleft}
Average performances measures, in percentages for PCALG, FFLDR and RIPE in the synthetic DAG of size $p = 100$ with different error structures. Numbers in parentheses indicate the standard deviation of the estimates over 50 draws of simulated data (only for PCALG and RIPE).
\end{flushleft}
\label{tbl:errcomp100}
\end{table}
\begin{table}[t]
\centering
\caption{{\bf Impact of increasing false positive and negative errors in perturbation data on estimation of high dimensional cyclic graphs.}}
{\footnotesize
\begin{tabular}{r|l|ccc|ccc|ccc|}
\multicolumn{1}{c}{\,} & \multicolumn{1}{c}{\,} & \multicolumn{3}{c}{$F_1$} & \multicolumn{3}{c}{$P$} & \multicolumn{3}{c}{$R$} \\ \cline{3-11}
\multicolumn{1}{c}{\,} & \multicolumn{1}{c|}{\,} & PCALG & FFLDR & RIPE & PCALG & FFLDR & RIPE & PCALG & FFLDR & RIPE \\ \cline{2-11} \cline{2-11}
\,      & NO ERR        & 27(3) & 58 & 63(1) & 31(3) & 89 & 83(1) & 25(3) & 43 & 50(1) \\ \cline{2-11}
\,      & $0.02\%$      & 27(3) & 47 & 63(1) & 31(3) & 63 & 83(1) & 25(3) & 38 & 51(1) \\
$FP$    & $0.04\%$      & 27(3) & 39 & 61(1) & 31(3) & 43 & 80(1) & 25(3) & 35 & 50(1) \\
\,      & $0.06\%$      & 27(3) & 34 & 59(1) & 31(3) & 33 & 73(1) & 25(3) & 35 & 50(1) \\ \cline{2-11}
\,      & $2\%$         & 27(3) & 57 & 61(1) & 31(3) & 88 & 83(1) & 25(3) & 43 & 49(1) \\
$FN$    & $4\%$         & 27(3) & 55 & 59(1) & 31(3) & 82 & 81(1) & 25(3) & 41 & 47(1) \\
\,      & $6\%$         & 27(3) & 54 & 59(1) & 31(3) & 82 & 81(1) & 25(3) & 40 & 46(1) \\ \cline{2-11}
\,      &$0.01\%$, $1\%$& 27(3) & 54 & 62(1) & 31(3) & 73 & 83(1) & 25(3) & 43 & 49(1) \\
$FP+FN$ &$0.02\%$, $2\%$& 27(3) & 48 & 61(1) & 31(3) & 64 & 80(1) & 25(3) & 39 & 50(1) \\
\,      &$0.03\%$, $3\%$& 27(3) & 42 & 61(1) & 31(3) & 51 & 79(1) & 25(3) & 35 & 50(1) \\ \cline{2-11} \cline{2-11}
\end{tabular}
}
\begin{flushleft}
Average performances measures, in percentages, for PCALG, FFLDR and RIPE in the synthetic cyclic network of size $p = 1000$ with different error structures. Numbers in parentheses indicate the standard deviation of the estimates over 5 draws of simulated data (only for PCALG and RIPE).
\end{flushleft}
\label{tbl:errcomp1000}
\end{table}
\subsection*{Regulatory Network in Yeast}\label{yeast}
To evaluate our method on real data, we explore the transcription factor (TF) regulatory network in \emph{Saccharomyces cerevisiae}. The influence matrix $\mathcal{P}$ was estimated from a large transcription factor knockout experiment \cite{Hu2007}, that was subsequently reanalyzed by \cite{Reimand2010}. Briefly, the experiment consists of 269 knockouts or knockdowns of yeast transcription factors, investigated by hybridization to microarrays under normal conditions. In total, 588 custom-made two-color microarrays with wild-type standard total RNA used as reference strain were employed in four batches and three different strains. Reanalysis of the data involved within-array normalization, background correction, between array-normalization and corrections for batch effects and strain effects. The genes were ranked using a moderated \emph{t}-statistic, and FDR-correction was applied based on the method of \cite{BenHoch:95}.

We extracted the expression values for the genes corresponding to the 269 TF knockouts (resulting in a square matrix of expression values) and a $p$-value cut-off was chosen to 0.002 based on the same type of plot as for the DREAM4 networks
(see Figure~S10). 
The resulting influence matrix corresponds to a graph with a strongly connected component of size 113.

The steady state experimental data employed come from a publicly available data set (ArrayExpress E-TABM-773) and has been used~(e.g., see~\cite{wageningen2010func}) to assess the day-to-day variation in large yeast array experiments. The data contains 200 samples, with samples from each day hybridized against a pool of wild-type strains. Although the samples are not necessarily independent, or identically distributed (due to batch effects, temporal correlations etc), here we use this dataset as an approximation for i.i.d measurements.
For the strong components whose size prohibits an exhaustive search, we apply the MC-DFS heuristic with $m=10,000$
random permutations.

\begin{figure}[!t]
 \begin{center}
\includegraphics[scale=0.55]{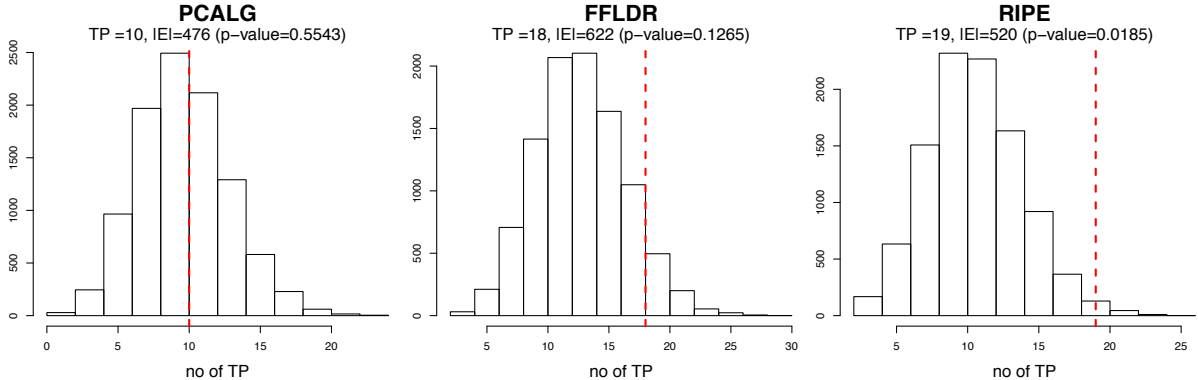}
\end{center}
\caption{
{\bf Performance evaluation for the reconstruction of the yeast regulatory network.} Number of true positives for each method, in comparison to the BIOGRID database, as well as a histogram for the number of true positives in randomly generated networks of the same size are shown. The p-values are obtained based on 10,000 randomly generated networks.}
\label{fig:yeastTP}
\end{figure}
The performance of the competing methods is based on the strategy outlined in the Preliminaries section above.
Figure~4 
shows the histogram of the number of true positives for randomly generated graphs of the same size as each of the reconstructed networks, as well as those obtained from three different methods
\footnote{The NEM algorithm failed to produce an estimate of the network in this case after 10 days of run time and was therefore not included.}.
Interestingly, the RIPE estimate is the only method for which the number of true positives are significantly larger (at 5\% significance level) than that obtained in a random graph (0.0185 for RIPE compared to 0.1265 for FFLDR and 0.5543 for PCALG). These findings further highlight the advantage of the proposed approach over existing methods.

A comparison of the estimated graphs using these three methods indicates a significant overlap between RIPE and FFLDR reconstructions: specifically, out of 520 edges detected in RIPE and 622 edges in FFLDR, 225 edges are in common. On the other hand, the PCALG reconstruction has considerable less overlap with the other two estimates: 8 and 12 common edges with RIPE and FFLDR, respectively. The significant overlap between the FFLDR and RIPE estimates suggests that some of the edges detected by both methods may indeed correspond to true regulatory interactions in yeast, which are not included in the BIOGRID database. Such results could be used as a starting point for designing the corresponding validation experiments.

As with the DREAM4 data, we also applied ARACNE for estimation of the yeast network, and found that the Bonferroni adjusted p-value cutoff of $1.38 \times 10^{-6}$ results in the best estimate, with 131 true positives (compared to BIOGRID) and 5594 total edges. This indicates that the estimated network based ARACNE is significantly denser compared to all other estimators. Evaluation of the significance of the number of true positives using the random graph method described above gives a $p$-value of $0.0286$, which indicates that ARACNE has a significantly larger proportion of true positives compared to FFLDR and PCALG. This is somewhat in agreement with the analysis on the DREAM4 data, where the combined performance of ARACNE based on the $F_1$ measure was affected by its low Recall rate compared to the other methods.

As pointed out in the Methods section, the RIPE algorithm can also be applied in the case where $k < p$. To illustrate this, we next describe the application of the RIPE algorithm for estimation of the entire regulatory network of yeast with $k = 269$ TF's and $p = 6051$ total genes. In such settings, one often obtains perturbation data on a subset of genes of interest (here the set of TFs) and is interested in obtaining an estimate of the \textit{regulatory interactions among the set of perturbed genes and all other genes} in the network. In this example, this amounts to a network with edges from each of the 269 TF's to each of the genes (TF's and TG's) in the network. Specifically, this corresponds to a 2-layer graph consisting of edges amongst TF's, as well as edges between TF's and TG's.

Considering the fact that perturbation data are only available for a subset of $k$ perturbed genes, one has to {\em impose} a constraint on the orderings between perturbed genes and the rest of the genes in the network. A natural constraint is to assume that no edges exist from unperturbed genes to the perturbed ones. This assumption defines a clear choice for ordering of nodes in the graph: the set of perturbed genes (say 1 to $k$) appear before the unperturbed ones in any ordering of nodes. It is then clear that RIPE can be applied to estimate the regulatory edges in the network by obtaining estimates of the two-layer network using the penalized regression approach in \eqref{DAGest_lasso_const}. The computational efficiency of the RIPE algorithms facilitates its application to estimation of the entire network regulatory interactions based on limited perturbation data. Estimation of the regulatory network of yeast with $p = 6051$ genes and $k = 269$ transcription factors based on 1000 orderings takes less than 22 minutes on a 2.7 GHz Laptop with 6 GB of memory. The resulting estimate includes 134 interactions reported in the BIOGRID dataset (true positives) and a total of 10014 edges. In an experiment similar to those reported above, the number of true positives in 1000 random graphs with the same number of edges and similar 2-layer structure no network with equal or larger true positives was observed ($p$-value $< 0.001$). The distribution of number of true positives in comparison to the number of true positives for the RIPE estimator are shown in Figure~S11.

\section*{Discussion}
The proposed methodology offers several advantages over existing approaches in addressing the key problem of reconstructing of regulatory networks. It relies on
a global assessment of causal orderings and employs both perturbation screens and steady state expression data for the reconstruction step that boosts performance. Further, the penalized likelihood method used for estimating the edges exhibits a certain degree of robustness to misspecification of the causal orderings, as observed in \cite{shojaie2010penalized}. As mentioned in the introductory section,
highly accurate perturbation data may be sufficient for the reconstruction task at hand, but this is not often the case. On the other hand,
integrating two data sources proves beneficial, as our numerical work illustrates.
We discuss next several issues related to the RIPE algorithm and outline some future research directions.

As previously indicated, scalability issues are important to the proposed methodology, since the influence matrix $\mathcal{P}$ usually contains cycles due to natural feedback loops in gene mechanisms and the noisy measurements in the perturbation experiments.
Hence, calculating \emph{all} possible causal orderings compatible with $\mathcal{P}$ may become infeasible given the exponential complexity of the problem. The proposed MC-DFS heuristic offers a fast, reliable alternative. Our numerical experiments suggest that in practice it is not required to exhaustively search the space of possible orderings, and a moderate number of randomly generated orderings often produce comparable estimates. Based on the results reported here, for graphs of up to $p=1000$ nodes, a total of $\sim 1000$ orderings results in reliable estimates.

Our extensive evaluation studies strongly suggest that the  RIPE algorithm is especially suitable in settings where one deals with noisy perturbation data and fairly good quality steady state
expression data are available. Algorithms utilizing only information from perturbation screens work well for topologies without many cycles, while those relying only on observational data
are not particularly competitive. Further, other data sources that can further filter the influence matrix, such as binding experiments (e.g. based on ChIP-chip technology), would be beneficial.

The proposed methodology is in principle applicable to other organisms for which steady state gene expression data with large sample size exist, such as Arabidopsis, mouse and human. The bottleneck would be the paucity of systematic perturbation screens, since they are more costly to produce. But the RIPE algorithm would be well suited for estimating a regulatory subnetwork for which adequate perturbation data are available.

An interesting extension of the proposed methodology involves perturbation screens from time course data \cite{frohlich2010fast}. The RIPE algorithm can be extended as follows. Relying on the fact that the bperturbation screens convey the causal ordering of the genes in the network, the DAG scoring method itself can be extended to cover time course steady state expression data. For example, a modified version of the likelihood scoring of DAGs can be used to shrink the estimated networks in each time point towards a common skeleton for the underlying network (similar to the approach described in \cite{guo2010joint}), or each time point can be modeled as a modified version of the network at the previous time point \cite{song2009keller}.

\section*{Acknowledgments}
The work was supported by NIH grants 1RC1CA145444-01 to GM and 1R21GM101719-01A1 to GM and AS.
The authors would like to thank Dr. Pinna for making available his code for the FFLDR method.

\bibliography{refs_v2}
\newpage

\end{document}